\documentclass[iicol, pdflatex,sn-mathphys-num]{sn-jnl}

\usepackage{graphicx}%
\usepackage{svg}
\usepackage{multirow}%
\usepackage{amsmath,amssymb,amsfonts}%
\usepackage{amsthm}%
\usepackage{mathrsfs}%
\usepackage[title]{appendix}%
\usepackage{xcolor}%
\usepackage{textcomp}%
\usepackage{manyfoot}%
\usepackage{booktabs}%
\usepackage{hyperref}
\usepackage{algorithm}%
\usepackage{algorithmicx}%
\usepackage{algpseudocode}%
\usepackage{listings}%
\usepackage{caption}
\usepackage{dblfloatfix}

\usepackage{csquotes}
\newcommand{\q}[1]{\enquote{#1}}

\theoremstyle{thmstyleone}%
\theoremstyle{thmstyletwo}%

\theoremstyle{thmstylethree}%

\begin{document}

\title[Article Title]{Enhancing Wide-Angle Image Using Narrow-Angle View of the Same Scene}


\author[]{\fnm{Hussain Md.} \sur{Safwan}}\email{safwan.du16@gmail.com}

\author[]{\fnm{Mahbub Islam} \sur{Mahim}}\email{mahbubislammahim@gmail.com}

\equalcont{These authors contributed equally to this work.}

\affil[]{\orgdiv{Department of Computer Science and Engineering}, \orgname{University of Dhaka}
}

\affil[]{\orgdiv{Department of Computer Science and Engineering}, \orgname{Jahangirnagar University}
}

\abstract{
A common dilemma in photography is the trade-off between capturing a wider field of view with less detail and a narrower field of view with finer detail. To address this, we propose a method that enhances wide-angle images by leveraging co-captured narrow-angle views of the same scene. Our approach introduces a narrow field-of-view (FoV) image-informed enhancement strategy that overcomes the limitations of blind super-resolution, which often struggles to recover high-frequency details without auxiliary guidance. We train a GAN-based model to extract fine-grained visual features from the narrow-view image and fuse them into the corresponding wide-view image through residual connections and an attention-based feature integration module. This fusion is guided by perceptual cues representing the visual essence—such as color, contrast, and sharpness—learned from the narrow FoV. We describe our architectural design and implementation in detail and demonstrate the effectiveness of our method through extensive experiments and comparisons with recent state-of-the-art techniques.
}




\maketitle

\section{Introduction}\label{sec1}
Modern smartphones and mirrorless cameras increasingly incorporate ultra-wide lenses \cite{ultrawidepatent} to capture expansive scenes in a single frame. While these lenses deliver a striking field of view, they often struggle to preserve the fine-grained detail achieved by conventional “primary” lenses. As discussed in the paper \cite{ultrawidedefects}, this difficulty is faced due to various technical issues, such as barrel distortion and spreading the view across a greater number of pixels, resulting in a lower perceived resolution. Users face a perennial trade-off: opting for a wider perspective at the expense of sharpness or retaining detail by narrowing the view. In this work, we seek to mitigate this dilemma by synthesizing wide-angle images whose perceptual quality matches that of narrow-angle shots, without sacrificing the original field of view. \\

While traditional super-resolution (SR) algorithms can increase the spatial resolution of an input image, they must rely on statistical priors and learned patterns to hallucinate high-frequency details, often leading to inconsistent results and visual artifacts. Our approach, on the other hand, hinges on jointly leveraging both narrow and wide-angle captures of the same scene. The narrow-angle image supplies high-resolution texture and detail, hence supplanting the blind dependence on learned distributions, while the wide-angle image carries context beyond the primary lens’s range. The enhancement objective, in this paper, is formulated as a resolution boosting task, informed by fine detail cues from the narrow FoV image, allowing us to draw upon proven strategies and validation procedures from SR literature. To that end, we introduce an end-to-end pipeline that utilizes cross-view attention scores to pinpoint the relevant features in the narrow FoV image embedding that can be used to guide the upscaling of the wide FoV counterpart. Inspired by the paper on Vision Transformer \cite{vit}, we adopt a patch-based approach to ensure the quality transfer occurs between regions that share visual similarity. \\

Central to our method is a proposed GAN \cite{gan} architecture: a generator attempting to fuse cross-view content and detail features through attention mechanisms and a discriminator that aims at enforcing both global realism and seamless patch blending. We further extend the classic Gram matrix representation to encode not only texture but also color, brightness, contrast, and sharpness metrics, ensuring rich visual fidelity. Applying this procedure iteratively across all available zoom levels, from telephoto to ultra-wide, our method aims to synthesize the broad perspective of the widest lens while improving detail quality using information from the primary lens. Our key contributions are,\\

\begin{itemize}
    \item An end-to-end pipeline that transfers high-resolution detail from narrow to wide-angle images.
    \item A GAN framework with cross-view attention and an enhanced Gram matrix formulation for robust detail propagation.
    \item A cascading lens-stack strategy that scales seamlessly from telephoto to ultra-wide captures.
\end{itemize}

\section{Related Works}\label{sec2}
Enhancing image quality has remained a central pursuit in computer vision. Earlier methods were primarily centered around super-resolution or upscaling techniques \cite{yang2010image}, which aimed to reconstruct high-resolution images from their low-resolution counterparts. While effective in producing sharper visuals, these models often introduced artifacts like oversmoothing, leading to a loss in texture and fine details. Later, generative adversarial models such as SRGAN \cite{ledig2017photo}, ESRGAN \cite{wang2018esrgan}, and Real-ESRGAN \cite{wang2021real} improved perceptual realism by emphasizing texture fidelity and producing visually convincing outputs. However, these models still struggle with preserving structural consistency.\\

Another stream of relevant research comes from neural style transfer. Initially developed for artistic applications \cite{gatys2016image}, these models demonstrated that image attributes could be transferred from one image to another. Subsequent work improved efficiency using feed-forward networks \cite{srcnn} and adaptive normalization techniques \cite{huang2017arbitrary}, broadening their use cases. However, transferring quality between narrow and wide FOV images introduces unique challenges not addressed by style transfer methods, as the wide FOV often contains additional contextual elements that require accurate integration.\\

\begin{figure*}[t]
    \centering
    \captionsetup{justification=centering}
    \includegraphics[width=\textwidth]{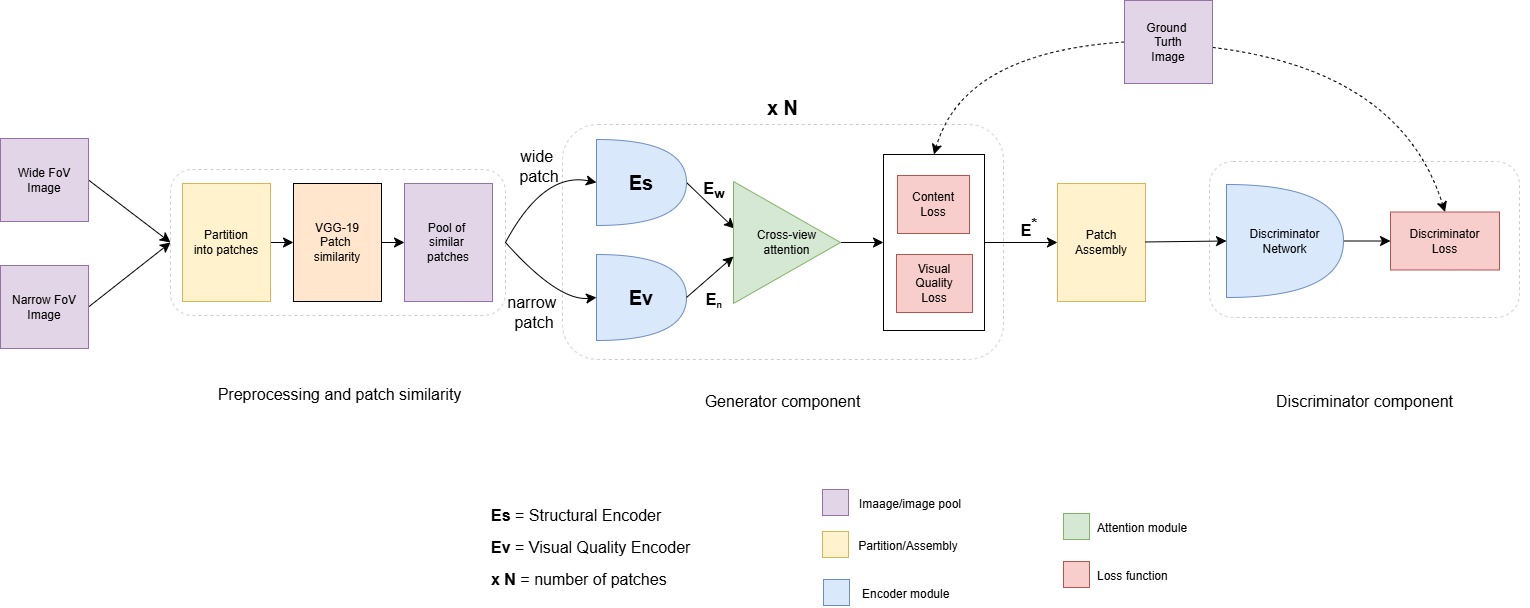}
    \caption{Overall model architecture}
    \label{fig:gan_arch}
\end{figure*}

Recent work includes LatticeNet \cite{latticenet}, which uses lattice-style blocks for adaptive feature fusion with low computation, enabling high-quality reconstruction without deep networks. However, its reliance on localized operations limits its capacity to capture long-range dependencies and global context, making it less effective in scenes with large structural variation or complex spatial arrangements. ELAN \cite{elan} combines shift convolutions with group-wise multiscale self-attention to capture both local and long-range dependencies, achieving a balance of accuracy and efficiency for real-time tasks. Nevertheless, the use of fixed attention windows introduces boundary discontinuities and restricts the model’s ability to fully capture cross-region interactions, particularly in non-uniform or geometrically diverse image regions.\\

Further exploring attention-based models, DLGSANet \cite{dlgsanet} employs a hybrid of Multi-Head Dynamic Local Self-Attention (MHDLSA) for local features and Sparse Global Self-Attention (SparseGSA) for key global cues, improving high-resolution reconstruction with lower compute. While this hybrid structure improves reconstruction quality, it models local features implicitly and may introduce irrelevant global tokens, leading to suboptimal detail preservation in fine-grained textures or highly cluttered scenes. Similarly, DMFFN \cite{dmffn} employs nested two-level asymmetric residual blocks to fuse multilevel features, enhancing edge preservation and accelerating convergence. However, its fixed fusion design limits adaptability across spatial scales, and deeper residual nesting can introduce training instability. These architectures, while advancing performance-efficiency trade-offs, may still face challenges in maintaining structural consistency under complex geometric variations.\\

To address these limitations, we propose a cross-view attention-based framework that explicitly models global context, aligns geometric structure across views, and adaptively fuses multi-scale features, leading to more robust reconstruction under challenging structural and visual conditions. By combining insights from generative models, reference-based super-resolution, attention mechanisms, and perceptual consistency techniques, our approach targets high-quality, structurally accurate image synthesis suited for immersive visual applications.

\section{Method}\label{sec3}
We define the task of perceptually enriching wide-angle imagery as a guided resolution enhancement problem. The model receives a pair of co-captured images: a wide field-of-view (FoV) image that offers spatial context, and a narrow FoV image that retains fine-grained visual fidelity. The output is a physically upscaled version of the wide FoV image, enriched with perceptual cues derived from the narrow FoV capture. While the model performs a spatial upscaling operation on the wide FoV input, the true goal is not just to increase pixel count, but to enrich the image perceptually. The resolution increase provides a larger canvas, enabling the injection of meaningful details such as sharp textures, balanced contrast, and natural colors. A detailed, step-by-step walk-through of the methodology is provided below. 

\subsection{Image Partitioning}
We start by dividing both the narrow and wide FoV images into fixed-sized patches to ensure portions of images sharing visual similarity are used to exchange content and visual quality data. Once partitioned, patches from wide and narrow FoV images that share visual properties are mapped using perceptual loss \cite{perceptualloss} and stored in the pool, ready to be pushed into the generator network. A cosine similarity on the feature maps generated by the mid layers of a VGG-19 \cite{vgg} backbone is used to determine the perceptual distance between patches. During inference, as further explained in the experimentation section, the same partitioning ratio is maintained.
\begin{figure}[H]
    \centering
    \includegraphics[width=1\linewidth]{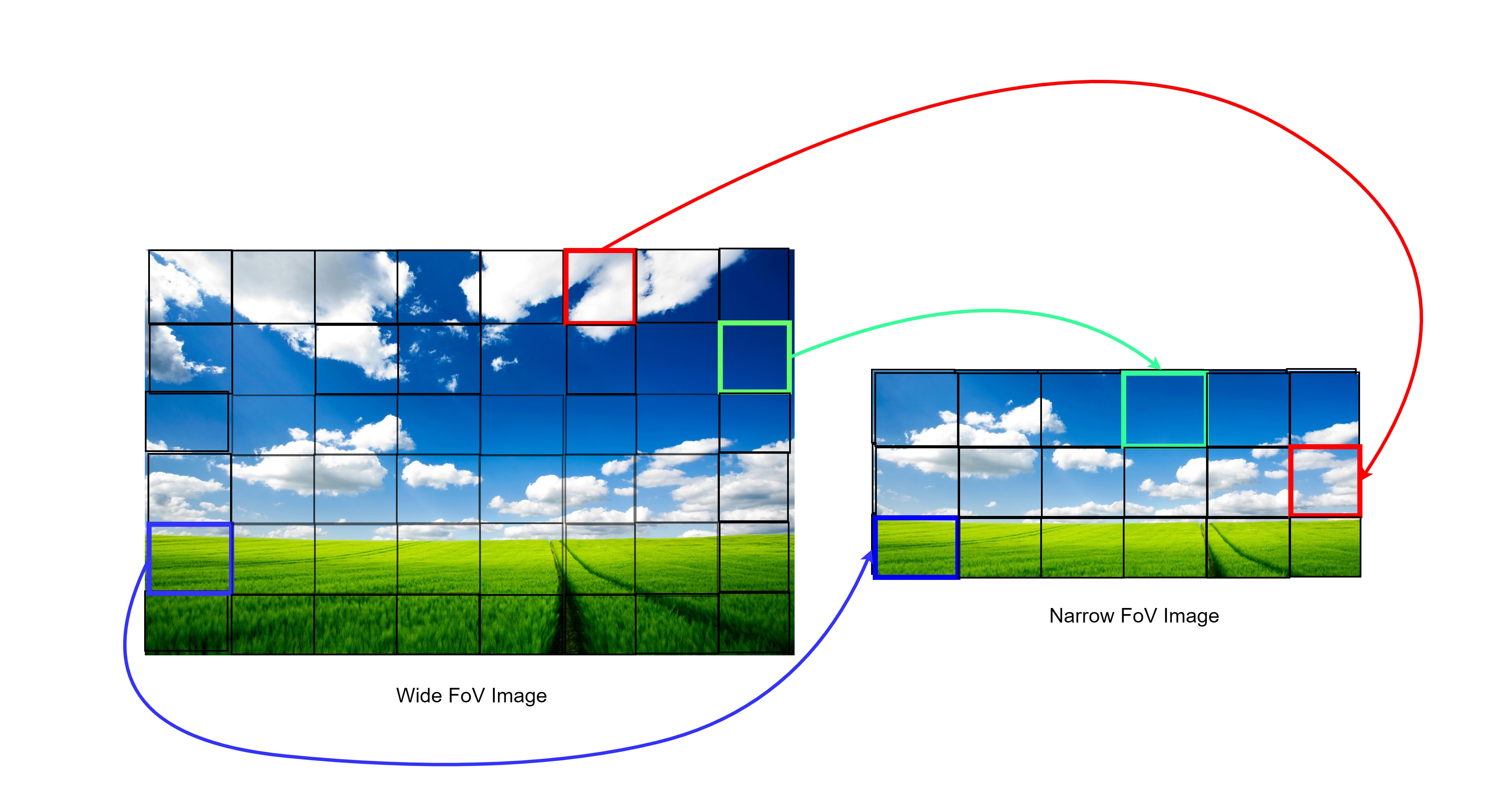}
    \caption{Image partitioning and patch similarity between narrow and wide FoV images.}
    \label{fig:patch}
\end{figure}

\subsection{Network Architecture}
The architecture of the generator network is built upon a stack of ResNet-like \cite{resnet} residual blocks that take as inputs a pair of wide and narrow FoV patches from the pool. The patches are encoded using separate encoder components to capture structural (content) and visual features from wide and narrow FoV patches, respectively. Each block then employs a cross-view attention mechanism to fuse the content and visual properties from wide and narrow feature maps, respectively. The pipeline consists of the following major components and has been visually illustrated in Fig \ref{fig:gan_arch}.

\subsubsection{Structural Feature Encoder} Wide FoV patches are encoded with traditional CNN-based encoders to capture the content details. Encoded wide patch represented as \(E_w\)

\begin{figure*}[t]
    \centering
    \includegraphics[width=1.0\textwidth]{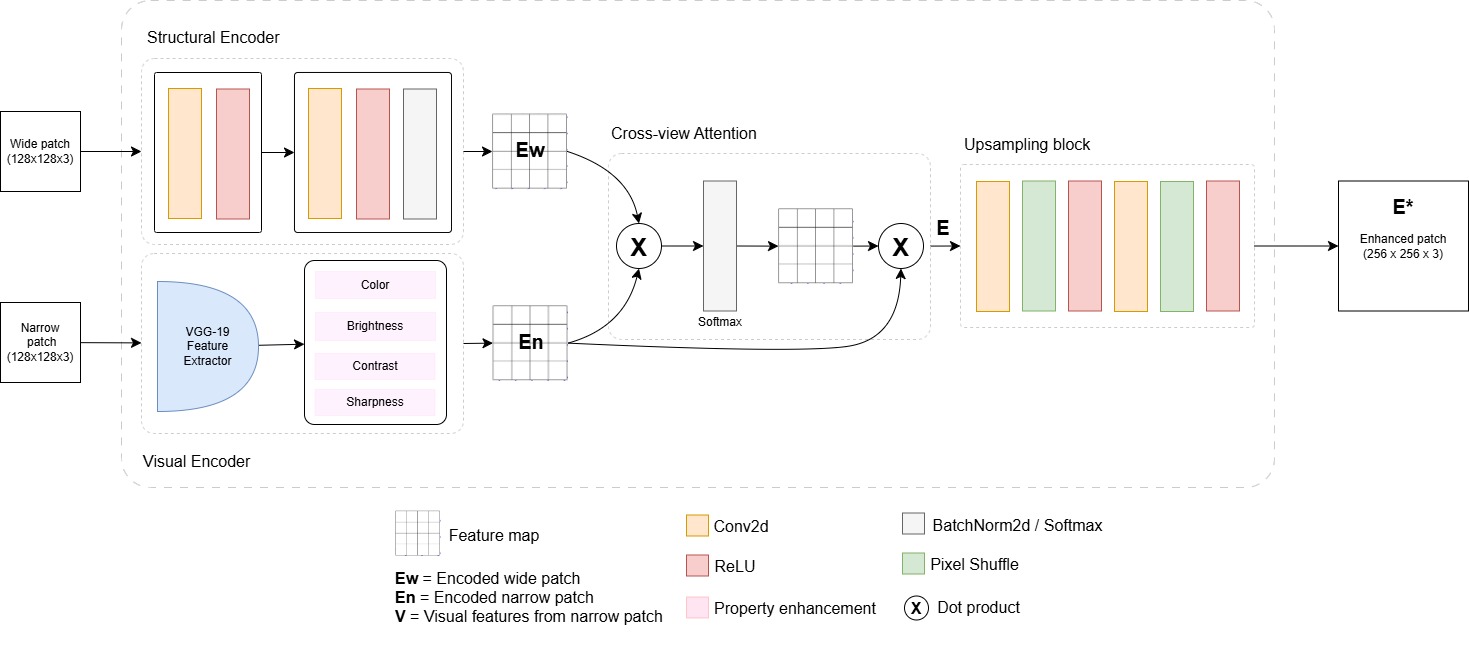}
    \caption{Detailed architecture for the Generator component}
    \label{fig:generator}
\end{figure*}

\subsubsection{Visual Feature Encoder} 
Narrow FoV patches are passed through encoders that are specially crafted to capture the visual essence of an image. We intend to do so by initially computing the Gram matrix representation of the patch and then augmenting it with visual cues computed from the patch. The augmented Gram matrix is then termed as \(E_n\). As mentioned earlier, we take the color, brightness, contrast, and sharpness parameters of a patch as the representative of its perceptual indicators. Detailed mathematical formulations are included in the supplementary material(Section 1). 

\subsubsection{Cross-Attention Residual Blocks}
Once the patches are encoded into the required latent space, we employ an attention-based fusion mechanism that fetches the most relevant visual features from the narrow patch and fuses those with the content features generated from the wide patch. This step facilitates the propagation of high-frequency details into the resultant image. 
\begin{itemize}
    \item Encoded wide FoV patch \(E_w\), representing the content data, acts as the query matrix
    \item Encoded narrow FoV patch \(E_n\), carrying the visual properties, provides both the key and value matrices. \(d_n\) is the dimension of the matrix \(E_n\)
    \item The attention scores, which gather semantically aligned, superior quality visual features present in the encoded narrow FoV patch, are computed in equation \eqref{eq_1}, using the scaled dot product equation introduced in the attention paper \cite{attention}, 
    \begin{equation}
    \label{eq_1}
        Attention (E_n, E_w) = softmax(\frac{E_w E_n^\top}{\sqrt{d_n}})E_n
    \end{equation}
    \item The gathered attention scores are then incorporated into the wide FoV feature map using a residual addition in equation \eqref{eq_2}, producing the feature map representation of the enhanced wide FoV patch, \(E\)
    \begin{equation}
    \label{eq_2}
        E = E_w + Attention (E_n, E_w)
    \end{equation}
\end{itemize}

\subsubsection{Upsampling Layers}
The output of the final \textbf{\textit{residual block}} is passed through a \textbf{\textit{pixel shuffle}} \cite{pixelshuffle} layer, spatially upscaling the patch to accommodate the incorporated perceptual cues, denoted by \(E^*\). Generator architecture has been illustrated in Fig \ref{fig:generator}

\subsection{Loss Calculation}
Generator loss is calculated patch-wise, i.e., each of the generated patches is compared against both the original wide and narrow FoV patches to quantify the content and visual deviations from the ground truth. Then the patches are assembled using seamless cloning to eliminate visible seams during blending, and the reconstructed image is passed through the discriminator, which computes a seam consistency loss over the entire image. The losses are described as follows, 

\subsubsection{Generator Loss}
\begin{enumerate}
    \item \textbf{Content Loss:} We find the deviations of the contents of the generated patch and ground truth with the following pixel-wise difference, 
    \begin{equation}
    \label{eq_3}
        L^{(l)}_{content}(E^o_w, E^g_w) = \frac{1}{n} \sum_{l}(E^o_w - E^g_w)^2
    \end{equation}
Here \(E^o_w\) and \(E^g_w\) are the latent space representations of the original and generated patches, and \(n\) is the dimension. 

    \item \textbf{Visual Quality Loss:} This loss is calculated as the weighted sum of differences of the Gram matrix representations of the generated patch and the narrow FoV patch, from mid layers of the VGG-19 backbone, using the following equation,
    \begin{equation}
    \label{eq_4}
        L^{(l)}_{visual} = \sum_l w_l ||E^o_n - E^g_n||^2
    \end{equation}
Here \(E^g_n\) and  \(E^o_n\) are the Gram matrix representations of the patches from the generated and the original narrow FoV patches, respectively. \(w_l\) is the weight assigned to the layer \(l\).

    \item Total generator loss can be written as,
    \begin{equation}
    \label{eq_5}
        L^{(l)}_{G} = L^{(l)}_{content} + L^{(l)}_{visual}
    \end{equation}
\end{enumerate}

\subsubsection{Discriminator Loss}
For the generated output image to be realistic and visually plausible, the results generated by the deconvolution network is passed through a discriminator component that aims to differentiate synthetic images from the real ones. We have designed two loss functions to assist the discriminator in this task. The losses are described as follows, 
\begin{enumerate}
    \item \textbf{Seam Consistency Loss}
This loss is aimed at eliminating the seams between the patches in the synthetic image by minimizing the perceptual difference at the boundaries of the adjacent patches. For adjacent patches \(P_i\) and \(P_j\), we take neighboring regions \(R_i\) and \(R_j\) which are \(b\) pixels wide and compute the loss as, 
\begin{equation}
    L^{(l)}_{seam} = \frac{1}{b}\sum ||\phi_l(R_j) - \phi_l(R_i)||^2
\end{equation}

    \item \textbf{Perceptual Loss against Ground Truth}
To ensure the essence of the narrow FoV has been plausibly incorporated into the synthetic image, the output is perceptually compared to the original wide FoV ground truth. We use the same VGG19 activation maps to find the perceptual difference \(L^{(l)}_{perceptual}\) between the generated wide FoV image and the original image from the dataset. 

Total discriminator loss is defined as follows, 
\begin{equation}
\label{eq_6}
    L^{(l)}_{D} = L^{(l)}_{seam} + L^{(l)}_{perceptual}
\end{equation}
\end{enumerate}

\subsection{Cascading Lens Stack}
The above procedure is incrementally applied across the available lens array to render the widest possible image with a resolution comparable to the narrowest available shot of the scene. For example, if a camera has lenses with the following zoom capacities: \(5x, 3x, 1x, 0.5x\); visual qualities are iteratively transferred from the \(5x\) to \(3x\), then from \(3x\) to \(1x\) and so on until the widest available shot has been reached.  

\section{Experimentation}
\subsection{Datasets}
We have used  \textbf{\textit{Landscapes Dataset}} \cite{lhq} (LHQ 1024×1024) to train the proposed model. With 90,000 high-quality images of 1024 x 1024 resolution, it includes a diverse range of urban and natural environments. These high-quality landscape images with rich textures and details provide a rigorous training ground for the model to learn how to handle complex natural scenes. We have randomly sampled \textit{20000 + 500} images (discussed in the \textit{Training Details} section) and fed them into the training loop in the ascending order of pixel variance \cite{scipy}, to ensure \q{easier} images are encountered first. Furthermore, to accommodate the rectangular aspect ratios of ultra-wide images encountered in real-world scenarios, we have fine-tuned the model on the  \textbf{\textit{DIV2K dataset}} \cite{Agustsson_2017_CVPR_Workshops}, which consists of 800 images at 2K (1920 × 1080; 16:9) aspect ratios, thereby facilitating the model’s ability to generalize effectively across varying aspect ratios.\\

\begin{figure}[t]
    \centering
    \includegraphics[width=0.75\linewidth]{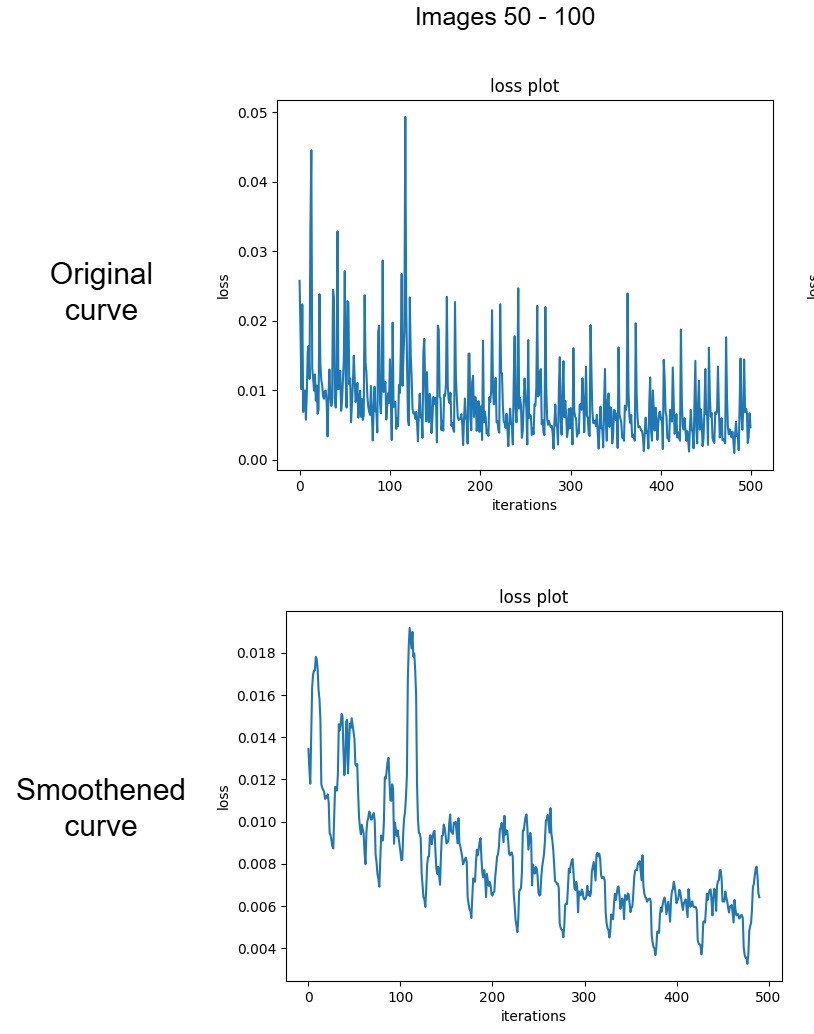}
    \caption{Convergence of loss curves shown over multiple iterations}
    \label{fig:loss}
\end{figure}

\begin{table*}[b]
    \centering
    \begin{threeparttable}[H]
    \centering
    \begin{tabular}{ccc}
         \textbf{Narrow Input} & \textbf{Wide Input} & \textbf{Output} \\
         \includegraphics[width=1.5in]{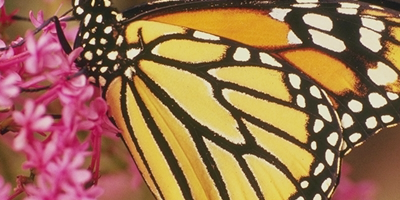} & 
         \includegraphics[width=1.5in]{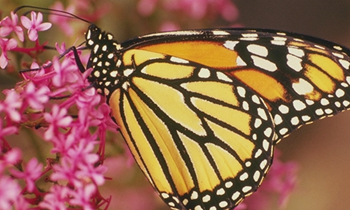} & 
         \includegraphics[width=1.5in]{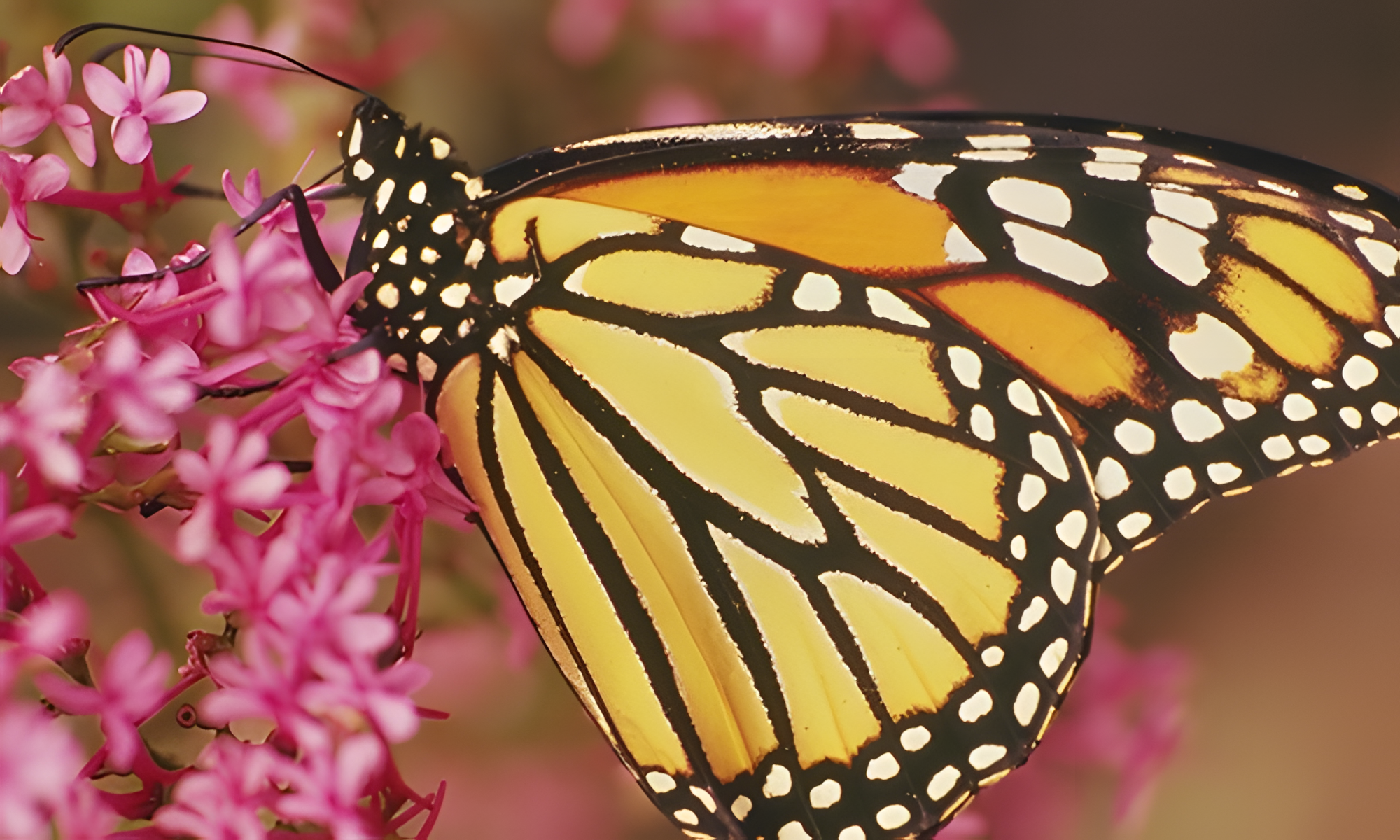} \\
         2x Zoom & 1x Zoom & 1x Output \\
         
         \includegraphics[width=1.5in]{primary_output.png} & 
         \includegraphics[width=1.5in]{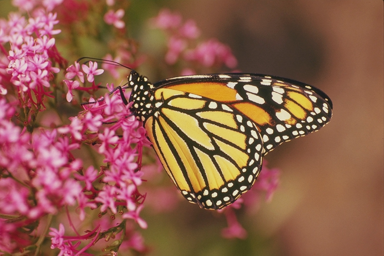} &
         \includegraphics[width=1.5in]{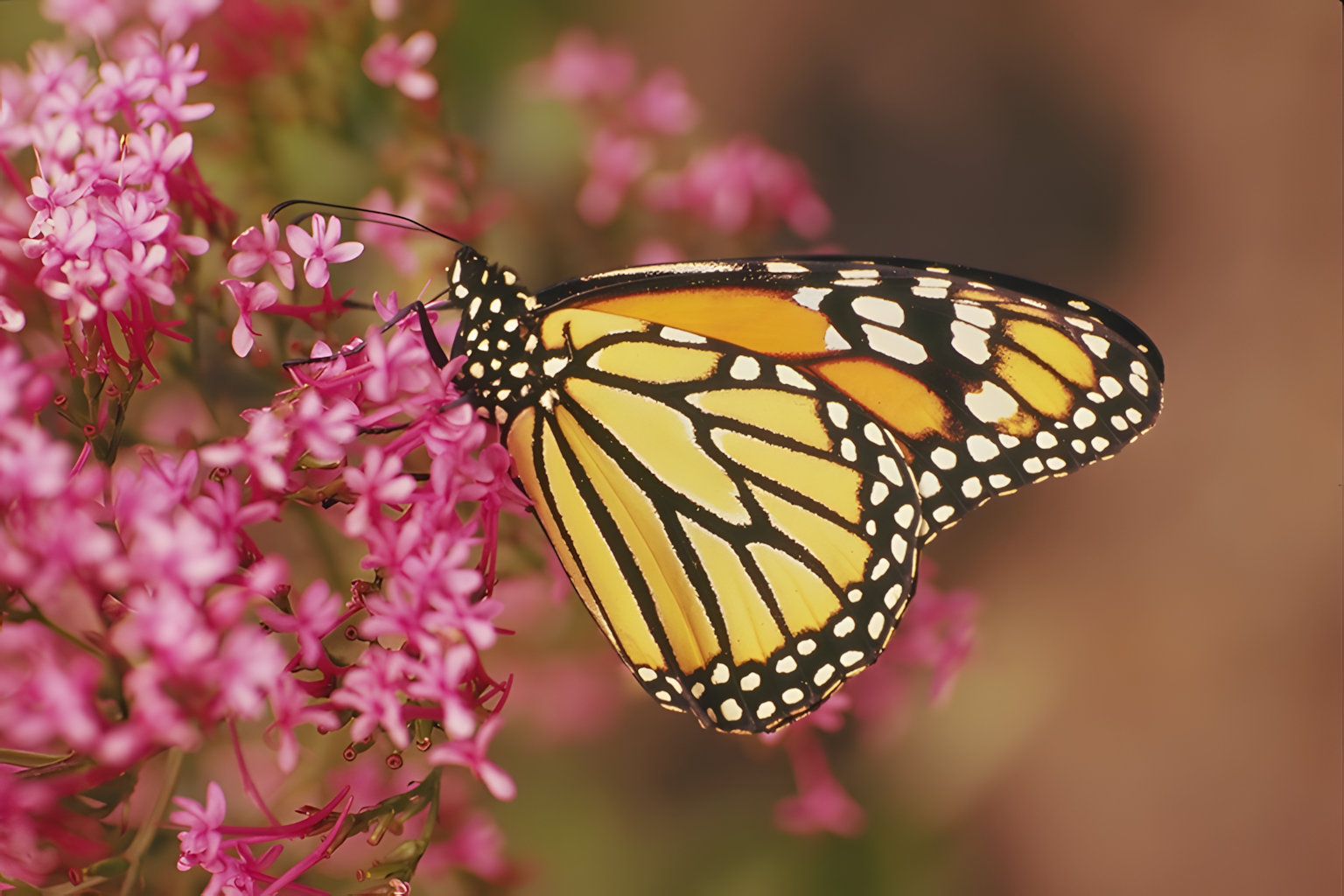} \\
         1x Output & 0.5x Zoom & 0.5x Output \\
    \end{tabular}
    \caption{Experimental results demonstrating the cascading effect of multi-zoom image upscaling.}
    \label{tab:cascading}
    \end{threeparttable}
\end{table*}

\subsection{Simulating Narrow and Wide FoVs}
High-quality narrow FoV simulation: For each image in the dataset, we extract a high-resolution central region with a zoom factor of \(\frac{5}{3}\). This region is then embedded into a larger blank canvas, preserving the original central region’s quality. The resulting image replicates the characteristics of a narrow FoV while maintaining the high quality of the original.\\

Low-quality wide FoV simulation:
We downsampled each image by applying a blur kernel and adding noise following this paper \cite{Zhang_2018_CVPR}, as well as resizing to 512 × 512, and then upsampled it (using bicubic interpolation) back to 1024 × 1024. This intentional degradation process mimics the wide image of low-quality imaging conditions while preserving the broader scene context, allowing us to test our methods under challenging visual conditions.

\subsection{Image Reconstruction Pipeline}
\begin{figure}[H]
    \centering
    \includegraphics[width=1.0\linewidth]{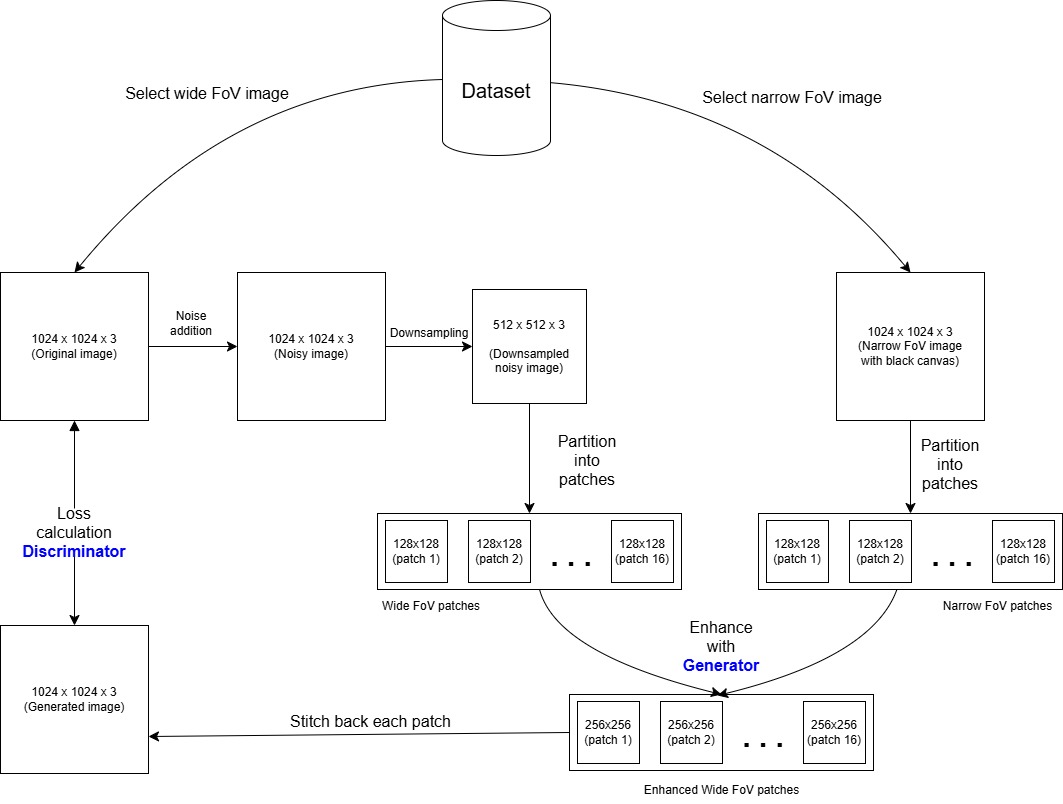}
    \caption{Enhanced image reconstruction pipeline}
    \label{fig:recons_pipeline}
\end{figure}

 During training, each image is passed through a reconstruction pipeline. The cycle begins with loading a pair of images, wide and narrow FoV, from the dataset, applying the preprocessing mentioned earlier, and dividing the processed images into patches. The patches, from the pool, are then fed into the generator, which produces their enhanced counterparts. The enhanced patches are stitched back into a single image, the super-resolved image, and forwarded to the discriminator for loss computation. This pipeline has been illustrated in Fig \ref{fig:recons_pipeline}. Implementation details about the generator and discriminator components are provided in the supplementary document (Section 2). 

\begin{figure*}
    \centering
    \includegraphics[width=0.9\linewidth]{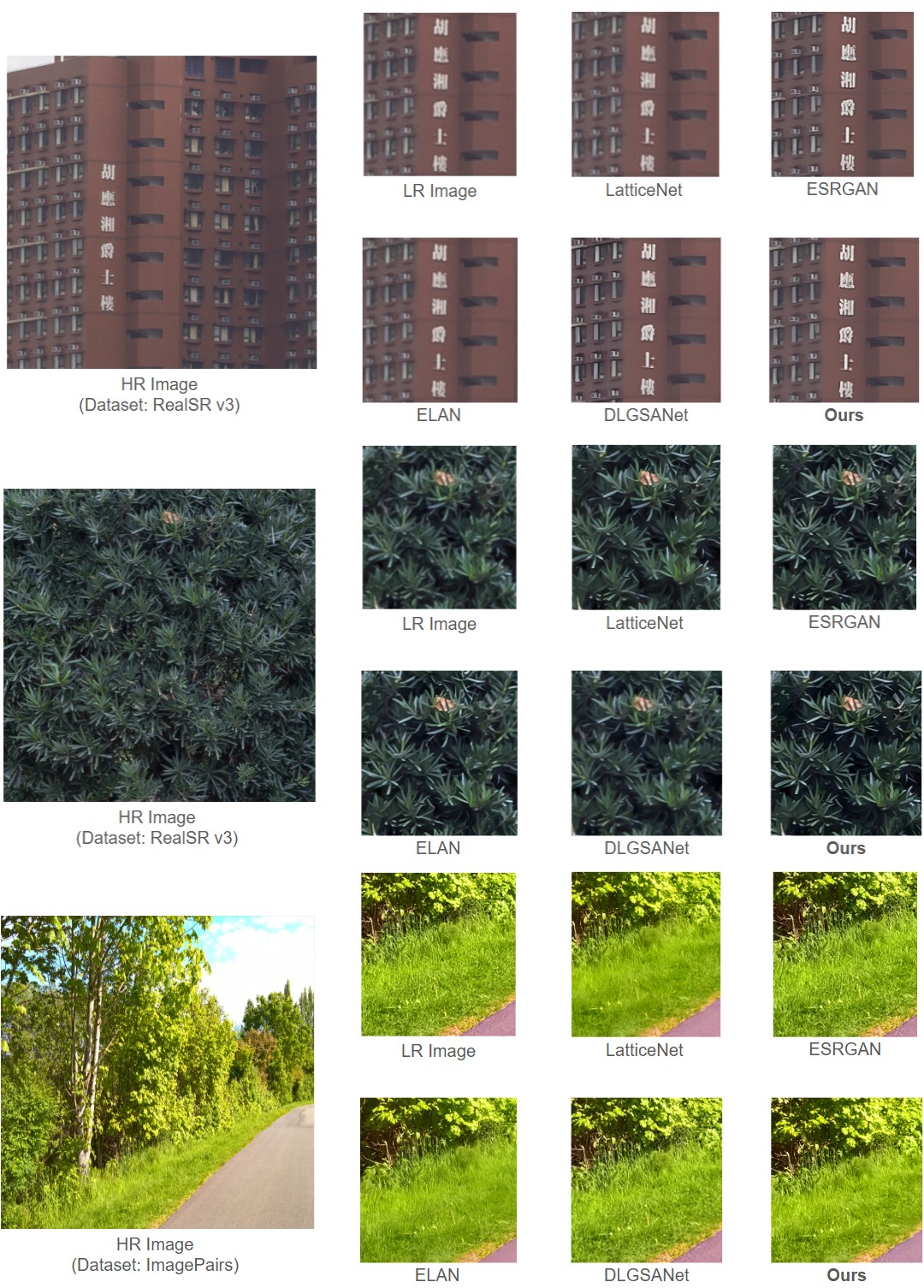}
    \caption{Visual comparison of our outcomes with existing methods.}
    \label{fig:visual_comp}
\end{figure*}

\begin{table*}[t]

    \centering
    \begin{threeparttable}
        \begin{tabular}{c|c|c|c|c|c|c}
            \textbf{Dataset} & \textbf{ESRGAN} & \textbf{LatticeNet} & \textbf{ELAN} & \textbf{DLGSANet} & \textbf{DMFFN} & \textbf{Ours}\\
            \hline
             \textbf{RealSRv3} & 0.2940 & 0.2907 & 0.2582 & 0.3409 & 0.3054 & \textcolor{red}{0.1740}\\
             \hline
             \textbf{ImagePairs} & 0.2979 & 0.2775 & 0.2754 & 0.2866 & 0.2872 & \textcolor{red}{0.2435}\\
             \hline
        \end{tabular}
        \caption{LPIPS Comparison (best marked in red)}
        \label{tab:lpips_comp}
    \end{threeparttable}

    \centering
    \begin{threeparttable}
        \begin{tabular}{c|c|c|c|c|c|c}
            \textbf{Dataset} & \textbf{ESRGAN} & \textbf{LatticeNet} & \textbf{ELAN} & \textbf{DLGSANet} & \textbf{DMFFN} & \textbf{Ours}\\
            \hline
             \textbf{RealSRv3} & 32.39 & 32.82 & 32.69 & 32.79 & 33.13 & \textcolor{red}{33.61}\\
             \hline
             \textbf{ImagePairs} & 29.21 & 29.23 & 29.13 & 29.09 & 29.09 & \textcolor{red}{29.29}\\
             \hline
        \end{tabular}
        \caption{PSNR Comparison (highest marked in red)}
        \label{tab:psnr_comp}
    \end{threeparttable}
    
    \centering
    \begin{threeparttable}
        \begin{tabular}{c|c|c|c|c|c|c}
            \textbf{Dataset} & \textbf{ESRGAN} & \textbf{LatticeNet} & \textbf{ELAN} & \textbf{DLGSANet} & \textbf{DMFFN} & \textbf{Ours}\\
            \hline
             \textbf{RealSRv3} & 0.8385 & 0.8479 & 0.8582 & 0.8433 & 0.8557 & \textcolor{red}{0.8870}\\
             \hline
             \textbf{ImagePairs} & 0.7244 & 0.7168 & 0.6782 & 0.7421 & \textcolor{red}{0.7604} & 0.7597\\
             \hline
        \end{tabular}
        \caption{SSIM Comparison (highest marked in red)}
        \label{tab:ssim_comp}
    \end{threeparttable}
    
\end{table*}

\subsection{Training Details}
Key hyperparameters for training and evaluation included an Adam optimizer with a learning rate of 0.0001 with exponential decay every 10 epochs, a batch size of 16, perceptual loss layers (Conv2\_2, Conv3\_2, Conv4\_2 of VGG19), and a threshold for patch similarity empirically set to 0.7. During training, an image is partitioned into 64 patches, and the patch dimensions are determined likewise, for example \(128\times128\) for Landscape dataset images. In case of inference, image dimensions are resized to the next multiple of 64, if needed, to facilitate the partitioning strategy.\\

The generator network was pre-trained for 50 epochs using content and visual losses before incorporating adversarial loss, each epoch carrying 10 sample images. The full generator-discriminator framework was then trained for an additional 2000 epochs, totaling 20.5K iterations. The convergence behavior of the generator loss has been illustrated in Fig \ref{fig:loss}. These curves represent the first 15 epochs where the generator has been trained solo. Upper row features the raw loss curves as generated by the GAN architecture, while the lower row has been smoothened using the \textbf{\textit{numpy convolve}} \cite{numpy} method.

\subsection{Experimental Setting}
For evaluation against existing works in the field, we have used 2 benchmark datasets which are \textbf{RealSRv3} \cite{realsrv3} and \textbf{ImagePairs} \cite{imagepairs}. These evaluation datasets differ from classical SR datasets in that the narrow and wide-FoV pairs are captured using actual focal length manipulation, as opposed to mathematical simulation, and hence are better suited to our particular case. We have, however, performed a parallel validation on traditional datasets \textbf{Set5} \cite{set5}, \textbf{Set14} \cite{set14}, and \textbf{BSD100} \cite{bsd100}, results of which are available in the supplementary document (Section 4). All reported metrics have been calculated based on the complete output image, as opposed to patch-wise measures. We have used the LR image provided by the dataset as the input, and the given HR as the ground truth.   

\subsection{Experiment Results}

Table \ref{tab:cascading} illustrates the cascading effect of the outputs derived from images captured at different zoom levels. In the first row, the input images consist of a 2x (telephoto) and a 1x (standard) zoom, which collectively produce a high-resolution 1x (standard) output. The second row takes the output from the previous row as one input and combines it with a low-resolution 0.5x (wide) image, ultimately generating a high-resolution 0.5x (wide) image. This sequence effectively simulates how our method can be applied in a cascading manner using all available zoom lenses. Visual outcomes on the benchmark datasets have been elaborated in the supplementary document (Section 3B).\\

Loss convergence in the epoch interval of 5 to 10 has been shown in Fig \ref{fig:loss}. The lower curve has been smoothed to denote the overall downward trend of the graph. Details on loss minimization has been included in the supplementary document (Section 3A).

\section{Comparison with Prior Arts}
We have evaluated our proposed methods on \textbf{LPIPS} \cite{lpips}, \textbf{PSNR}, and \textbf{SSIM} metrics over datasets mentioned earlier. For all the metrics, we have compared our results with existing technologies such as \textbf{\textit{ESRGAN}} \cite{wang2018esrgan}, \textbf{\textit{LatticeNet}} \cite{latticenet},  \textbf{\textit{ELAN}} \cite{elan}, \textbf{\textit{DLGSANet}} \cite{dlgsanet} and \textbf{\textit{DMFFN}} \cite{dmffn} and presented the results in Tables \ref{tab:lpips_comp}, \ref{tab:psnr_comp}, and \ref{tab:ssim_comp} respectively. We have also included a visual comparison against the prior methods in the Fig \ref{fig:visual_comp}.

\subsection{Quantitative Analysis} 
In terms of perceptual quality, as assessed using the LPIPS metric, our approach shows a clear improvement over earlier methods. This is largely due to the integration of visually similar patches, the selection of representative perceptual features, and the inclusion of perceptual loss within the discriminator, all of which contribute to more visually realistic outputs.

Similarly, our model achieves a noticeably higher PSNR score compared to other techniques. Notably, on the ImagePairs dataset, it is the only method to surpass a PSNR value of 30.

In the case of SSIM, our method also achieves a marked advantage on the RealSRv3 dataset, indicating superior preservation of structural details. However, for the ImagePairs dataset, the DMFFN model slightly outperforms ours. This can likely be attributed to its Multi-stage Feature Supplementation, which enhances geometrical consistency by combining shallow and deep features—albeit with increased computational demands.

\subsection{Qualitative Analysis}
We evaluated 2× super-resolved outputs from the referenced datasets against those produced by existing models. First, many of these models lack broader content-aware guidance, making it difficult for them to accurately reconstruct structural details—this is especially evident in elements like building windows (first image). Second, earlier methods tend to produce less distinct edges, often resulting in soft or blurred boundaries, whereas our approach delivers noticeably sharper edges, as illustrated by the leaf contours in the second image. Lastly, in visually rich scenes with high color saturation and contrast—such as green grass (third image)—previous models often fail to preserve vibrancy and fine texture. In contrast, our method successfully maintains these visual attributes by utilizing cues from the narrow field-of-view input, which is rich in detail.

\section{Conclusion}
This paper introduces an approach for enhancing ultra-wide field-of-view (FoV) images by leveraging paired narrow-FoV views of the same scene. The core of our method is a patch-based cross-view attention mechanism that selectively synthesizes fine-grained details from narrow-FoV patches into their wide-FoV counterparts, guided by a Gram matrix-based visual encoding to capture the texture statistics for accurate alignment. To further refine the reconstruction, we introduce a cascading lens stack mechanism that progressively transfers details across adjacent lens images, facilitating spatial consistency and coherent integration across the full image. \\

The model trained is on the Landscape HQ dataset to cover both structural diversity and finer details, and fine-tuned on DIV2K for handling rectangular images. For evaluation, we have used standard benchmark datasets including RealSRv3 and ImagePairs. Experimental results demonstrate a promising improvement over strong baselines such as ELAN, DLGSANet, DMFFN, and LatticeNet, especially in LPIPS, along with both PSNR and SSIM metrics. Qualitative comparisons further add to the perceptual superiority of our results. \\

Limitations to our current work include the necessity of further fine-tuning our model with real-life ultra-wide and narrow FoV pairs, since it is trained on synthetic data. Also, additional optimizations are required to deploy the model on-device to serve real-time purposes and video super-resolution tasks.  
\\

\section*{Declarations}
\begin{itemize}
    \item \textbf{Funding:} This research did not receive external funding.
    \item \textbf{Competing Interests:} The authors do not have any competing interests as defined by Springer, or other interests that might be perceived to influence the results and/or discussion reported in this paper.
\end{itemize}

\bibliography{sn-bibliography}

\end{document}